\documentclass{article}
\usepackage{spconf,amsmath,graphicx}
 \usepackage{xcolor}
  
   \usepackage{graphicx}
\usepackage{enumitem}
\usepackage{amsmath,amsfonts}
\setlist[enumerate]{itemsep=0pt,parsep=0pt,topsep=0pt}
\usepackage{hyperref}
\usepackage{float}

\newcommand{\blue}{\color{blue}}
\title{Contrastive Learning for Regression on Hyperspectral Data}
\name{Mohamad Dhaini\textsuperscript{1,2 *}, Maxime Berar\textsuperscript{1}, Paul Honeine\textsuperscript{1}, Antonin Van Exem\textsuperscript{2} \thanks{* Corresponding Author \newline This work is funded by Tellux Company and ANRT (Association Nationale de la Recherche et de la Technologie).}}

\address{\textsuperscript{1} 
Univ Rouen Normandie, INSA Rouen Normandie, Université Le Havre Normandie, \\
Normandie Univ, LITIS UR 4108, F-76000 Rouen, France
\\
\textsuperscript{2} Tellux, 76650 Petit-Couronne, France}

\begin{document}
%
\maketitle
\begin{abstract}
Contrastive learning has demonstrated great effectiveness in representation learning especially for image classification tasks. However, there is still a shortage in the studies targeting regression tasks, and more specifically applications on hyperspectral data. In this paper, we propose a contrastive learning framework for the regression tasks for hyperspectral data. To this end, we provide a collection of transformations relevant for augmenting hyperspectral data, and investigate contrastive learning for regression. Experiments on synthetic and real hyperspectral datasets show that the proposed framework and transformations significantly improve the performance of regression models, achieving better scores than other state-of-the-art transformations.

\end{abstract}
\section{Introduction}
\label{sec:intro}
Hyperspectral imagery offers valuable insights into the physical properties of an object or area without the need for physical contact. Hyperspectral sensors capture a broad range of information within the light spectrum, often spanning hundreds of contiguous bands across a wavelength range of approximately 500 nm to 2500 nm. This capability allows each material to possess its own distinct spectral signature. This type of data has been gaining significant attention from the signal processing and machine learning community, and became a direct application for classification \cite{roy2019hybridsn}, regression \cite{niu2021deep}, unmixing \cite{dhaini2022end} and object detection  \cite{yan2021object} tasks.

Recently, due to the limitation of supervised learning techniques to labeled data,
self-supervised learning methods have been gaining  popularity to learn general representations from unlabeled data thus having more discriminative features in the used neural networks. In this context, contrastive learning \cite{chen2020simple} is a self-supervised approach that provides such discriminative features by maximizing the similarity between similar data examples and minimizing it between dissimilar ones. The use of such approach on hyperspectral data is still encountering some challenges especially regarding the augmentation techniques to be used. Data augmentation techniques often used for general images (e.g., image rotation) are not applicable to hyperspectral data. 
%
In this article, we investigate the use of contrastive learning to improve  regression results on hyperspectral data, with application in hyperspectral unmixing and pollution estimation. The contributions can be seen as following:
\begin{enumerate}
    \item We revisit some popular augmentation techniques, often used in computer vision, to fit into hyperspectral data. Besides, we make use of well-known radiative transfer models that simulate the atmospheric effect on hyperspectral data to generate augmented spectra.
    \item We adapt the cross-entropy based contrastive loss to fit regression tasks by incorporating the use of a ball of given radius to select positive and negative pairs.
    \item We show the performance of our method in real-world scenario, including unmixing and prediction of pollution concentration in soil data.
\end{enumerate}
The rest of the paper is organized as follows. Section \ref{sec:related} highlights some of the related work on contrastive learning with hyperspectral data. In Section \ref{sec:method}, we present the core ideas behind our proposed method. The experimental studies with the obtained results are presented in Section \ref{sec:results}. The contributions and future steps are summarized in Section \ref{sec:conclusion}.

\section{Related Work}
\label{sec:related}

Recently, there have been some studies investigating the use of contrastive learning on hyperspectral data. However, the majority of these studies were targeting classification tasks.  \cite{wang2023nearest} introduced a contrastive learning network based on a nearest neighbor augmentation scheme, 
by extracting similarities from nearest neighbor samples to learn enhanced semantic relationships. 
In \cite{li2023unlocking}, 
three data augmentation methods were introduced to enhance the representation of features extracted by contrastive learning. These methods were band erasure, gradient mask and random occlusion. In the same context, \cite{hu2021deep} introduced 
a spectral-spatial contrastive clustering network, with 
a set of spectral-spatial augmentation techniques that includes random cropping, resizing, rotation, flipping, and blurring for the spatial domain, as well as band permutation and band erasure for the spectral domain. \cite{ou2022hyperspectral} introduced a framework for detecting surface changes in hyperspectral images using self-supervised learning. Their main contributions involved an augmentation technique based on a Gaussian noise and a contrastive loss based on Pearson coefficient and negative cosine correlation. Similarly, \cite{cai2022superpixel}
introduced a 
neighborhood contrastive subspace clustering network for unsupervised classification of large hyperspectral images based on a superpixel pooling autoencoder. 
Besides, \cite{braham2022self} 
demonstrated that by pre-training an encoder on unlabeled pixels using the proposed 
Barlow-Twins algorithm, accurate models can be obtained even with a small number of labeled samples. For Unmixing, the authors in \cite{hernandez2023contrastive} introduces in addition to the standard reconstruction error loss often used, a contrastive loss is applied to the endmember matrix to promote separability between endmembers and another regularization loss to encourage the minimum
simplex volume constraint in endmembers.

\section{Method}
\label{sec:method}

\subsection{Proposed Framework}

\begin{figure}
    \centering
    \includegraphics[width=.49\textwidth]{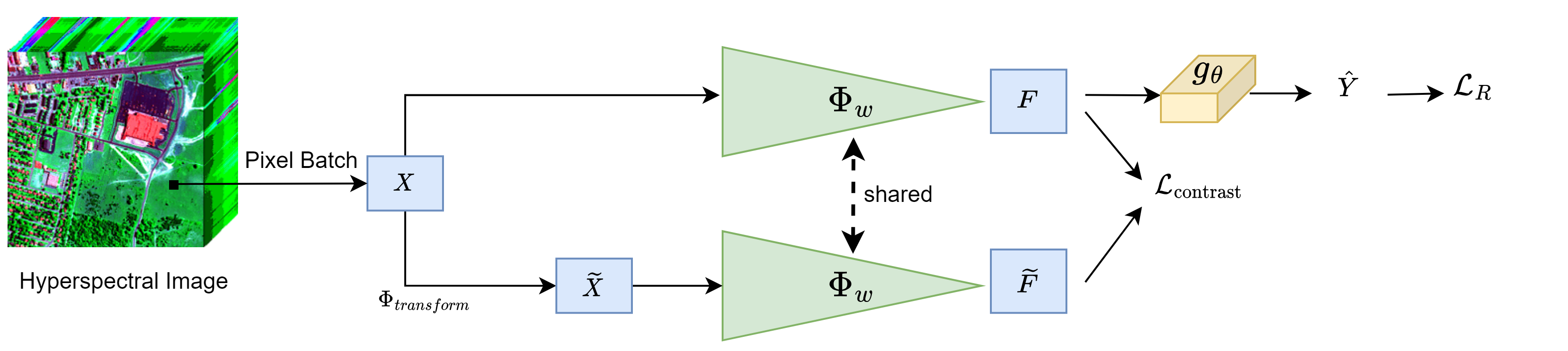}
    \caption{Architecture of the proposed method.}
    \label{fig:archi}
\end{figure}

The task we are addressing is a pixel-level regression on hyperspectral data. The process starts by extracting a batch of $N$ pixels $X = [x^1(\lambda),x^2(\lambda),\ldots,x^{N}(\lambda)]^\top \in \mathbb{R}^{N \times b}$ from a hyperspectral image, where  $\lambda$ is the wavelength with $b$ total number of wavelengths. $X$  is then transformed using a defined spectral transformation $\Phi_{transform}$ to get $\widetilde{X}$, which will be passed along with the original batch to a shared feature extractor $\Phi_w$ to get $\widetilde{F}$ and $F = [f^1,f^2,\ldots,f^{N}]^\top$ respectively. These features are passed into a regression network $g_\theta$ that will generate the regression labels $\hat{Y} = [\hat{y}^1,\hat{y}^2,\ldots,\hat{y}^{N}]^\top \in \mathbb{R}^{N \times s}$ for $s$ prediction variables. The network will be trained with a joint contrastive and regression losses simultaneously. The architecture of the proposed method can be seen \autoref{fig:archi}.

\subsection{Spectral Data Augmentation Methods}

Most of data augmentation techniques  focus on the spatial domain of images such as geometric transformations, noise injection, and color distortions. These techniques cannot be adapted directly to the spectral domain as the used transformations should not create strong deformations to the original spectrum in order not to loose useful information. In the following, we propose a new list of transformations that can be adequate for the spectral domain:


\begin{enumerate}
\item The Spectral Shift involves shifting the spectrum in the wavelength, such as
\begin{equation}
\label{eq:spectral_shift}
\widetilde{x}(\lambda) = x(\lambda - \Delta),
\end{equation}
 By applying random shifts, the model can learn to be more robust to spectral variations in the input data.

\item The Spectral Flipping involves reversing the order of spectral bands in a spectrum according to the following:
\begin{equation}
\widetilde{x}(\lambda) = {x}(\lambda_{min} + \lambda_{max} - \lambda).
\end{equation}
where $\lambda_{min}$ and $\lambda_{max}$ are the minimum and maximum wavelengths in the spectrum, respectively. This augmentation technique can help the model learn the invariance to the order of spectral bands and handle potential inconsistencies in the spectral ordering across different datasets or sensors.

\item The Scattering Hapke's Model \cite{hapke1981bidirectional} simulates realistic scattering effects caused by light interaction with surfaces according to the following formula:
\begin{equation}
\widetilde{x} = \frac{{\omega}}{{(1+2\mu_1\sqrt{1-\omega})(1+2\mu_2\sqrt{1-\omega})}},
\end{equation}
\begin{equation}
    w =1 -\left(\frac{\sqrt{\mu_0^{4}{x^2} + (1+4\mu_0^2 {x})(1-x)} - 2\mu_0x} {1+4\mu_0^2x}\right)^2
\end{equation}
where $\omega$ is the single scattering albedo of the material, and $\mu_1$ (resp. $\mu_2$ ) is the cosine of the angle between the incoming (resp. outgoing) radiation and the normal to the surface and $\mu_0$ is the initial cosine angle of the incoming radiation. By varying $\mu_1$ and $\mu_2$, we can generate spectra with different scattering effect. 

\item The Atmospheric  Compensation model \cite{uezato2016novel} helps the model learn to handle atmospheric effects. Assuming full visibility and that the adjacency effect is negligible, this model is given by:
\begin{equation}
    \widetilde{x}=x  \frac{E_{\text {sun-gr }} \mu_1+E_{\text {sky }}}{E_{\text {sun-gr }} \mu_2+E_{\text {sky }}},
\end{equation}
where $E_{\mathrm{sun}-\mathrm{gr}}$ denotes the solar radiance observed at the ground level, and $E_{\mathrm{sky}}$ denotes the skylight. The parameters $\mu_1$ and $\mu_2$ are the cosines of the angles between the surface normal and the direction of the sun at each pixel and at the calibration panel, respectively. 

\item The Elastic Distortion consists in a displacement grid on the wavelength axis, such as
\begin{equation}
\widetilde{x}(\lambda) = {x}( \lambda  + \epsilon(\lambda)) = {x} \Big( \lambda  + \sum_{i=1}^{N_G} A_i e^{-\frac{(\lambda-\lambda_i)^2}{2\sigma^2}}\Big)
\end{equation}
where $\epsilon(\lambda)$ is the random displacement function applied to the input signal,  $N_G$ is the number of Gaussian kernels used to generate the distortion,  $A_i$ is the amplitude of the $i$-th Gaussian kernel, $\lambda_i$ 
its center wavelength
, and $\sigma$ controls its width
. This technique can help the model learn to handle spectral variations caused by distortions or misalignments.
\end{enumerate}

In addition to this proposed list, there exist some augmentation techniques that are already presented in literature:
\begin{enumerate}

\item The Band Erasure \cite{hu2021deep} randomly removing certain wavelength from the spectral data. 

\item The Band Permutation \cite{hu2021deep} involves randomly permuting the order of the spectral bands.

\item  The Nearest Neighbor \cite{wang2023nearest} involves creating new synthetic samples based on the average of the closest samples. 

\end{enumerate}

\subsection{Contrastive Learning}

After augmenting the data ${X}$ with different views $\widetilde{X}$, both are forwarded to the feature extractor $\Phi_w$, which extracts the corresponding features, ${F}$ and ${\widetilde{F}}$  respectively, and then will be later on passed to a regression head $g_{\theta}$. To optimize the feature extractor, we train the neural network so that two spectral data transformed via different augmentation techniques and 
with close regression labels should share similar features in the latent space, while different spectra with different labels should be far away. This type of training can be achieved using a self-supervised contrastive loss \cite{chen2020simple}. 

In classification, selecting similar data pairs (referred to as positive pairs) can be done by taking the transformed version of an image as well as other images that belong to the same class, while dealing with the rest as negative ones. In regression, as we do not have class labels, the alternative is to define for the $i$-th sample a ball $\mathbf{B}^i$ of radius $r$ where positive pair $j$ is selected as following:
\begin{equation}
r \geq \left \| y^i-y^j\right \|_2.
\end{equation}

The common contrastive loss used in most recent work is based on the cross entropy, which can be written as
\begin{equation}
\small
\mathcal{L}_{Contrastive}=-\frac{1}{N} \sum_{i= 1}^{2N} \sum_{j \in \mathbf{B}^i} \log \frac{\exp \left(\operatorname{sim}\left(f^i, f^j\right) / \tau\right)}{\sum_{k \not \in \mathbf{B}^i}  \exp \left(\operatorname{sim}\left(f^i, f^k\right) / \tau\right)}
\label{hard-contrast}
\end{equation}
where $\operatorname{sim}(\boldsymbol{u}, \boldsymbol{v})=$ $\boldsymbol{u}^T \boldsymbol{v} /(\|\boldsymbol{u}\|\|\boldsymbol{v}\|)$ is the cosine similarity between two vectors, and $\tau$ is a temperature scalar. By minimizing this loss, the similarity between samples $i$-{th} and $j$-{th} is maximized while minimizing the similarity between $i$-{th} and $k$-{th} samples. For training, the contrastive loss is combined with a standard mean squared error regression loss according to the following:
\begin{equation}
     \mathcal{L}_{total}=\mathcal{L}_{\text{R}}+ \alpha \ \mathcal{L}_{\text{Contrastive}},
\end{equation}
\begin{equation}
     \mathcal{L}_{\text{R}}=\frac{1}{N} \sum_{i=1}^{N} \left\| {y}^{i} -
     g_{\theta} \left(f^i\right) \right\|^{2}.
\label{loss_regress}
\end{equation}

\section{Expermiments \& Results}
\label{sec:results}

\subsection{{Synthetic Data}}
For the synthetic data, four random endmembers were selected from the 
USGS digital spectral library \cite{swayze1993us}. Each endmember is composed of 224 contiguous bands. A total of $100 \times 100$ pixels were generated with abundances following a Dirichlet distribution. Additive zero-mean Gaussian noise was added to the data with a signal-to-noise ratio of 20 dB. We considered a polynomial post-nonlinear mixing of the endmembers where the nonlinearity is represented by the element-wise~product of two linear mixtures as following:
\begin{equation}
x=M a+Ma\odot Ma + n,
\label{pnmm}
\end{equation}
with $M$,$a$ and $n$ being endmember matrix, abundances and noise vector respectively and $\odot$ denotes the element-wise~product. The model was trained to predict the abundances from the input of mixed spectra. For the architecture of $\Phi_w$, we used three fully connected layers mapping the input shape to 128, 64 and 32 nodes respectively. For the regression network $g_\theta$, we use two fully connected layers that further reduces the dimension to 16 then to $s=4$ abundances to be estimated in this case. 
We trained our network with 100 epochs and a batch size of 32. We evaluated the influence of each transformation with two regression metrics, $R^2$ score and the mean absolute error (MAE). The results are shown in \autoref{tab:synthetic}. We can see that the all the transformations presented in the table improved the results compared to the baseline model where no contrastive loss was applied. Besides, we can see that the shift and elastic transformations provided the top two results. These can be justified as elastic transform can create similar adjustments to spectral data to that created by the shift transform. Besides, in \autoref{fig:error}(a) we compare  the prediction error distribution with and without the use of the contrastive loss. We can see that the mean of the error distribution is shifted closer to zero due to the contrastive learning, which highlights the improvement in the prediction values.

\subsection{Real Soil Data} 

\begin{figure}
    \centering
    \includegraphics[height=5cm,
    width=0.49\textwidth]{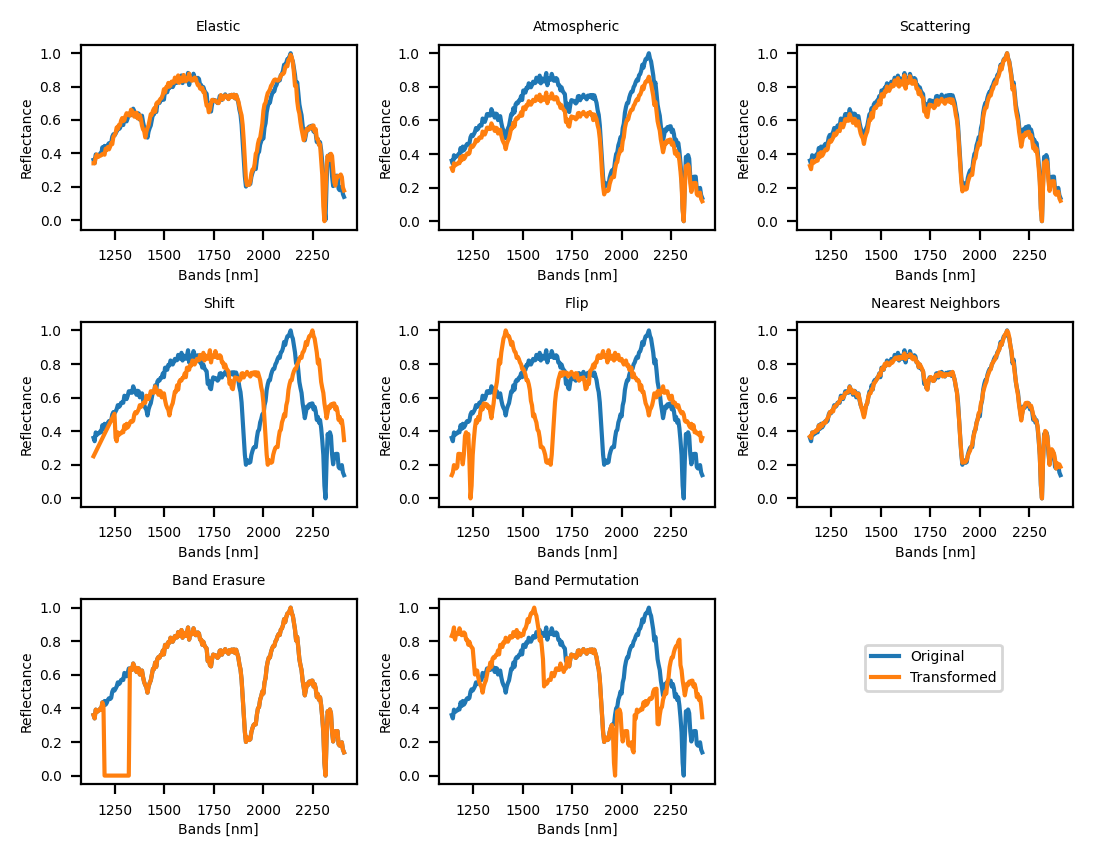}
    \caption{Original and Transformed Examples from Real Data.}
    \label{fig:tellux-data}
\end{figure}

\begin{figure}
    \centering
    \includegraphics[
    width=.5\textwidth]{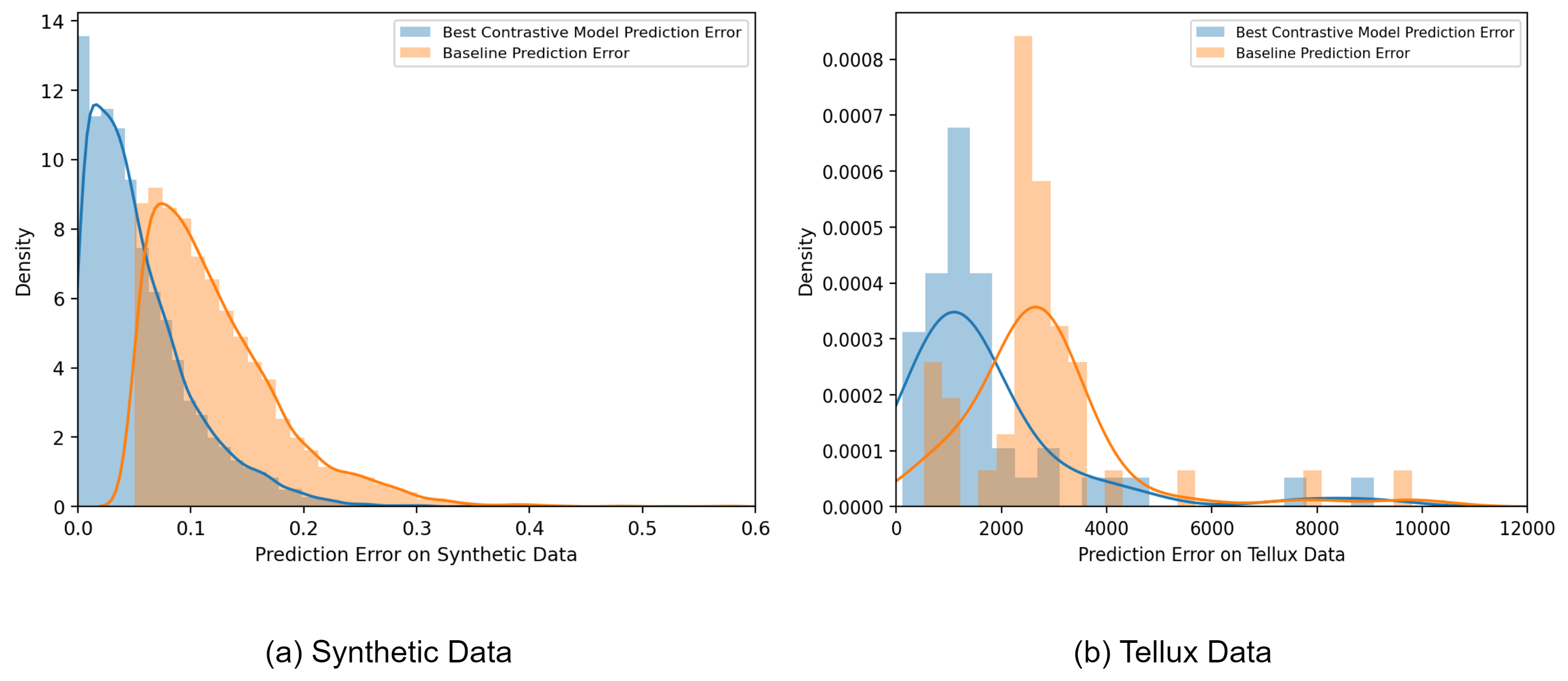}
    \caption{Prediction Error Distribution.}
    \label{fig:error}
\end{figure}

To evaluate our model on a more challenging environment, we considered a dataset provided by 
Tellux 
for soil pollution analysis using hyperspectral imaging \cite{dhaini2021hyperspectral}. 
The used soil is formed of a mix between different soil matrices (sand, clay, silt, organic material ...). The dataset is composed 10000 spectra with a spectral range [1130-2450 nm] and containing hydrocarbon pollution concentration ranging [0-20000 mg/kg] where each soil mixture is mixed with hydrocarbon pollutants. An example of these spectral data as well as their transformed versions obtained by the proposed transformations can be seen in  \autoref{fig:tellux-data}.
We used the same settings (architecture, batch size, ...) as for the synthetic data and trained it with 1000 epochs. In the same way as with the synthetic data, 
the obtained results given in \autoref{tab:tellux} show a clear improvement in $R^2$ and MAE scores for all models with contrastive loss compared to the baseline model. Besides, shift, flip and elastic transformations provided the best results, which can be justified as these methods give invariance to the model for spectral bands order while maintaining the spectral profile. In addition, \autoref{fig:error}(b) shows the shift in the distribution error thanks to the use of the contrastive loss.

\vspace{-1em}
\begin{table}[H]
    \caption{{Regression Results on Synthetic Data.
    }}
    \centering
    \small
    \resizebox{.4\textwidth}{!}
    {
    \begin{tabular}{lrr}
& $R^2$\qquad\qquad & $\mathrm{MAE} ~\quad$ \\
\hline Baseline (No Contrastive) & $0.55 \pm 0.004$ & $0.073\pm0.07$ \\
\hline Band erasure \cite{hu2021deep} & $0.62\pm0.003$ & $0.064\pm0.05$ \\
\hline Band Permutation \cite{hu2021deep} & $0.63\pm0.003$ & $0.063\pm0.05$ \\
\hline Nearest Neighbor \cite{wang2023nearest} & $0.61\pm0.004$ & $0.065\pm0.05$ \\
\hline Scattering & $0.64\pm0.004$ & $0.061\pm0.06$ \\
\hline Atmospheric & $0.65\pm0.004$ & $0.059\pm0.06$ \\
\hline Flipping & $0.62\pm0.005$ & $0.062\pm0.07$ \\
\hline Elastic & ${0 . 6 6   \pm 0 . 0 0 3}$ & ${0 . 0 5 8 \pm 0 . 0 5}$ \\
\hline Shift & $ {\blue{{0 . 7 5\pm 0 . 0 0 3}}}$ & $\blue{{0 . 0 5 3 \pm 0 . 0 5}}$ \\
\hline
\end{tabular}}
    \label{tab:synthetic}
\end{table}

\vspace{-2.5em}
\begin{table}[H]
    \caption{Regression Results on Real Data.}
    \centering
    \small
    \resizebox{.4\textwidth}{!}
    {
    \begin{tabular}{lrr}
& $R^2$\qquad\qquad & $\mathrm{MAE} ~\quad$ \\
\hline Baseline (No Contrastive) & $0.45\pm0.002$ & $2274.04\pm21.2$ \\
\hline Band erasure \cite{hu2021deep} & $0.54\pm0.003$ & $1620.22\pm18.1$ \\
\hline Band Permutation \cite{hu2021deep} & $0.53\pm0.003$ & $1700.04\pm20.5$ \\
\hline Nearest Neighbor \cite{wang2023nearest} & $0.54\pm0.003$ & $1850.40\pm34.5$ \\
\hline Scattering & $0.56\pm0.003$ & $1737.26\pm18.2$ \\
\hline Atmospheric & $0.55\pm0.002$ & $1796.63\pm16.1$ \\
\hline Elastic & $0.58\pm0.002$ & $1709.33\pm18.3$ \\
\hline Flip & $0.59\pm0.002$ & $1708.52\pm18.4$ \\
\hline Shift & $\blue{{0.59\pm0.002}}$ & $\blue{{1380.37\pm16.5}} $ \\
\hline
\end{tabular}}
    \label{tab:tellux}
\end{table}

\subsection{Combination Study}
To provide more robustness of the regression model to various types of variability that might be present in the spectral data, we combine several spectral transformations from the ones presented before. 
To reduce the number of all possible combinations from the eight presented transformations, we propose an incremental procedure where we start by taking the transformation that provided the best result (namely the shift transformation from \autoref{tab:synthetic} and \autoref{tab:tellux}) and then we do all the 2-element combinations. After selecting the best pair, we repeat the process to select the third transformation, and so on. 
\autoref{tab:ablation1} and \autoref{tab:ablation2} provide the $R^2$ scores of the regression model, as well as the difference $\Delta R^2$ for each incremental update. For simplification, we only show the incremental settings that led to the best combinations.  We can see that combining the shift, atmospheric, elastic and scattering transforms can improve the regression metrics when combined together.

\begin{table}[H]
    \caption{Combination Study Results on Synthetic Data.}
    \centering
    \small
    \resizebox{.4\textwidth}{!}
    {
    \begin{tabular}{lcc}
 & $R^2$ & $\Delta R^2 $ \\
\hline Shift  & $0.7522$ & $-$ \\
\hline Shift + Atmospheric & $0.7774$ & $0.0252$ \\
\hline Shift + Atmospheric + Scattering & $0.7921$ & $0.0147$ \\
\hline Shift + Atmospheric + Scattering + Elastic & $0.7922$ & $0.0001$ \\
\hline
\end{tabular}}
    \label{tab:ablation1}
\end{table}

\begin{table}[H]
    \caption{Combination Study Results on Real Data.}
    \centering
    \small
    \resizebox{.4\textwidth}{!}
    {
    \begin{tabular}{lcc}
 & $R^2$ & $\Delta R^2$ \\
\hline Shift  & $0.59000$ & $-$ \\
\hline Shift + Atmospheric & $0.60639$ & $0.01639$ \\
\hline Shift + Atmospheric + Elastic & $0.61791$ & $0.01152$ \\
\hline Shift + Atmospheric + Elastic + Scattering & $0.61793$ & $0.00002$ \\
\hline
\end{tabular}}
    \label{tab:ablation2}
\end{table}

\section{Conclusion}
\label{sec:conclusion}
In this paper, we investigated the ability of using contrastive learning for regression tasks on hyperspectral data. We presented a set of spectral transformations adequate for hyperspectral data
. Besides, a contrastive loss was added to the training and a clear improvement was seen on the results of both synthetic and real datasets. Future work involves combining the presented framework with another domain adaptation frameworks to generalize knowledge on unseen domains.

\vfill\pagebreak

\end{document}